\documentclass[11pt]{article}

\usepackage[final]{acl}   
\usepackage{times}
\usepackage{latexsym}
\usepackage[T1]{fontenc}
\usepackage[utf8]{inputenc}
\usepackage{microtype}
\usepackage{graphicx}
\usepackage{booktabs}
\usepackage{amsmath}
\usepackage{amssymb}
\usepackage{multirow}
\usepackage{xcolor}
\usepackage{tikz}
\usetikzlibrary{positioning, arrows.meta, backgrounds, calc, decorations.pathreplacing}
\usepackage{CJKutf8}  

\title{Qwen-MT-Lumen at WMT2026: Instruction-Specialized Translation
via Quality-Aware Training and On-Policy Distillation}

\author{
\begin{tabular}{c}
\normalfont Hao Wang \quad Yangyang Liu \quad Xiaohu Zhao \quad Heng Liu \quad Zifu Shang \\[1pt]
\normalfont Tianhao Li \quad Ruize Gao \quad Jialong Tang \quad Liangying Shao \quad Haoran Wei \\[1pt]
\normalfont Baosong Yang \quad Linlong Xu \quad Longyue Wang \quad Weihua Luo \\[3pt]
\normalfont{Alibaba Token Hub} \\[3pt]
\normalfont\texttt{huaiyu.wh@alibaba-inc.com}
\end{tabular}
}

\begin{document}
\maketitle

\begin{abstract}
We present Lumen\footnote{The open-source model is available at
\url{https://huggingface.co/ATH-MaaS/Qwen-MT-Lumen}.}, our submission to the
WMT 2026 General Machine Translation Shared Task.
We focus on instruction-specialized translation, where systems must produce
accurate translations while satisfying fine-grained requirements on style,
terminology, and structured output.
Lumen follows a four-stage pipeline.
First, continued pre-training and supervised fine-tuning jointly strengthen multilingual translation competence and the ability to follow translation-specific instructions, such as terminology, style, and format constraints.
Second, we extend the originally offline M\textsuperscript{2}PO preference-optimization framework into an online preference-optimization stage: the current policy samples candidate translations, an LLM judge assigns continuous direct-assessment scores, and the resulting multi-pair preferences are used to update the model.
Third, on-policy distillation transfers stronger translation-instruction-following behavior from a larger teacher model to the Qwen3-14B student, improving translation-instruction adherence while retaining a 14B primary translation backbone.
Finally, structure-aware inference and postprocessing preserve required
formats, while selective repair addresses clear structural or decoding
failures before submission.
We participate in 23 translation directions.
Under an absolute $1$--$5$ LLM-judge rubric that evaluates translation-instruction following and translation quality separately, Lumen reaches an overall mean score of $4.60$. Despite using a 14B translation backbone, it achieves performance comparable to leading proprietary systems in both translation quality and translation-instruction following.
\end{abstract}

\section{Introduction}
\label{sec:intro}

The Conference on Machine Translation (WMT) continues to serve as the
primary arena for benchmarking progress in machine translation
\citep{kocmi2024wmt24,kocmi2025wmt25}. For years the field was dominated by the
Transformer architecture \citep{vaswani2017attention}, whose powerful
attention mechanism set a high standard for translation quality. The
recent advent of large language models (LLMs), however, has sparked a
paradigm shift: pre-trained on vast amounts of text, these models
exhibit remarkable abilities in understanding context, generating
fluent text, and leveraging world knowledge, making them exceptionally
promising foundations for translation systems
\citep{hendy2023how,alves2024tower}. This shift is already evident at
recent WMT campaigns, where many of the strongest general-translation
submissions are now built on LLM foundations
\citep{wang2025marco,zheng2025hunyuan,zhang2025systran}.

Despite the strong performance of general-purpose LLMs on translation,
adapting them into high-fidelity, specialized translation systems
remains challenging. Existing LLM-based MT
systems largely focus on generic translation quality, whereas in
practical applications users often require the model to follow
fine-grained instructions during translation: specifying a translation
style (e.g., formal, colloquial, or customer-service tone), adhering to
domain-specific terminology, resolving ambiguity from context, or
preserving the integrity of structured formats such as JSON or
rich text. General translation models struggle to satisfy such
requirements reliably, which constitutes a key gap for deploying
LLM-based MT in real-world scenarios. To bridge this gap, we propose
training and optimization methods for \emph{instruction-specialized
translation} that systematically improve both translation quality and
instruction-following ability.

This paper describes Lumen, our submission to the WMT2026
General Machine Translation Shared Task, covering 23 translation
directions and built on the Qwen3 foundation \citep{yang2025qwen3}. The
system consists of a four-stage pipeline, illustrated in
Figure~\ref{fig:pipeline}. Our main contributions are as follows:

\begin{itemize}
  \item \textbf{An instruction-specialized two-phase adaptation
  strategy.} We first run continued pre-training (CPT) on a corpus
  deliberately assembled to follow the WMT domains and language
  directions, mixed with general instruction data, which pulls the
  backbone towards the evaluation distribution while preserving its
  general ability. On top of this CPT-strengthened backbone we design an
  SFT recipe that mixes general translation
  data with translation-specific instruction data. A high proportion of
  translation data preserves the base translation ability, while
  instruction data covering the four WMT26 instruction categories
  (formal/informal voice, glossaries, structured translation, and style
  and expressions) enables the model to
  internalize instruction-following ability during fine-tuning. The
  instruction data is built cost-effectively through an LLM-assisted
  synthesis pipeline.

  \item \textbf{An online preference optimization framework based on
  M\textsuperscript{2}PO.} We extend the originally offline
  M\textsuperscript{2}PO
  (Multi-Perspective Multi-Pair Preference Optimization)
  \citep{m2po2026} into an online preference-optimization stage, coupling
  on-policy sampling with an LLM-judge (direct-assessment) reward. Unlike
  an offline procedure over a fixed preference set, this framework refreshes
  candidate translations from the current policy and converts their ranked
  scores into multi-pair supervision. The remaining gap in instruction
  compliance is addressed by the subsequent OPD stage.

  \item \textbf{An OPD recipe for translation instruction following.} We
  adapt on-policy distillation to instruction-specialized translation. After
  CPT+SFT+RL, our 14B model is strong at translation but remains weaker at
  following fine-grained translation instructions. We therefore use an
  off-the-shelf, open-source Qwen3-32B-Instruct model as teacher and our
  14B model as student, aligning token-level distributions on trajectories
  sampled by the student. This transfers the teacher's stronger
  instruction-following behavior to the student while preserving the
  student's translation specialization and its 14B primary translation
  backbone
  \citep{kim2016sequence,gu2024minillm}.

  \item \textbf{Structure-aware inference and selective repair.} At
  inference time, Lumen processes plain text and structured inputs in a
  format-aware manner, preserving markup and protected content through
  postprocessing. Rule-based checks identify clear structural or decoding
  failures, for which the system applies deterministic repair when
  possible and selectively rewrites the remaining cases.
\end{itemize}

On the WMT26 test sets, our 14B system attains an overall mean of $4.60$
under an absolute $1$--$5$ LLM-judge rubric (translation-instruction
following $4.49$, translation quality $4.71$). Despite using a 14B
translation backbone, Lumen achieves performance comparable to leading
proprietary systems in both translation quality and
translation-instruction following. The remainder of
this paper is organized as follows:
Section~\ref{sec:system} describes the four-stage system;
Section~\ref{sec:experiments} presents the
experimental setup and results; and Section~\ref{sec:conclusion}
concludes.

\section{System Description}
\label{sec:system}

\subsection{System Overview}
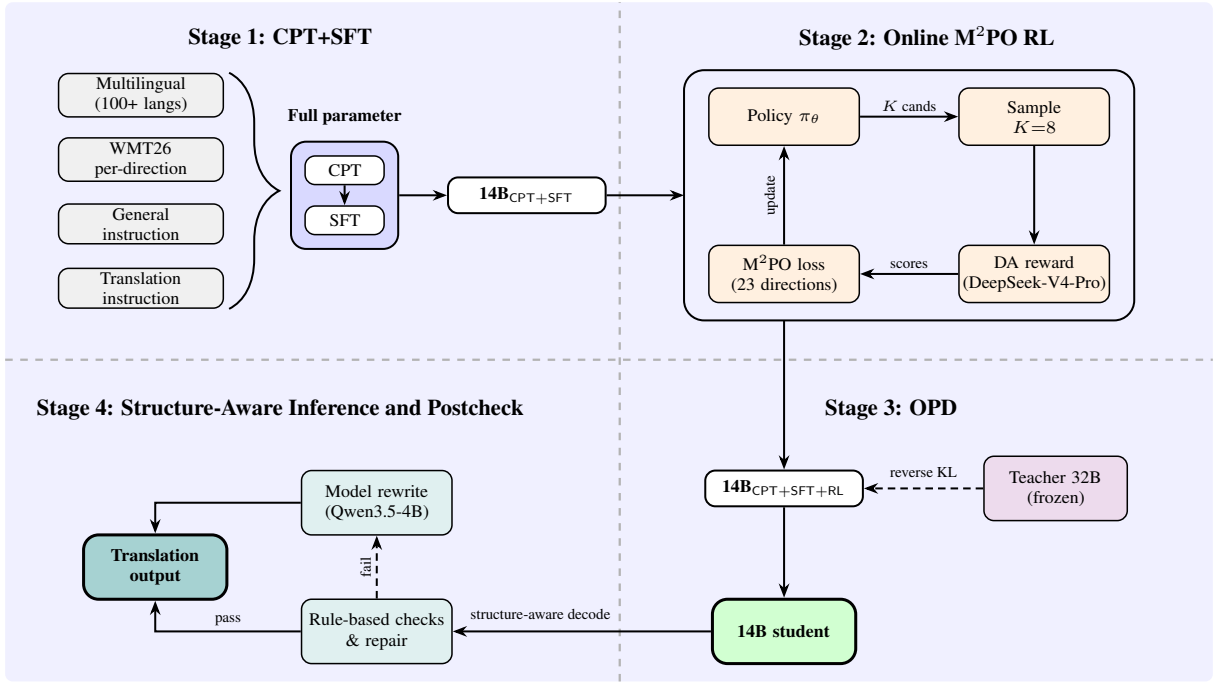
\begin{figure*}[!t]
  \centering
  \resizebox{\textwidth}{!}{%
  \begin{tikzpicture}[
    font=\small,
    >={Stealth[length=2mm,width=1.5mm]},
    every path/.style={thick, line width=0.6pt},
    data/.style   ={draw, rounded corners=4pt, fill=gray!12,
                    text width=22mm, minimum height=5mm, align=center,
                    font=\scriptsize, inner sep=1.5pt},
    bigbox/.style ={draw, thick, rounded corners=6pt, fill=white,
                    align=center, font=\small\bfseries, inner sep=3pt},
    submod/.style ={draw, rounded corners=4pt, fill=white,
                    text width=20mm, minimum height=6mm, align=center,
                    font=\scriptsize, inner sep=1.5pt},
    model/.style  ={draw, thick, rounded corners=4pt, fill=white,
                    text width=21mm, minimum height=5mm, align=center,
                    font=\scriptsize\bfseries, inner sep=1.5pt},
    final/.style  ={draw, very thick, rounded corners=6pt, fill=green!18,
                    text width=19mm, minimum height=6mm, align=center,
                    font=\scriptsize\bfseries, inner sep=1.5pt},
    ar/.style     ={->, thick},
    ardash/.style ={->, thick, densely dashed},
    edgelbl/.style={font=\tiny, midway, above=-0.6mm, sloped},
    stagelbl/.style={font=\footnotesize\bfseries, align=center},
  ]

    \begin{scope}[on background layer]
      \fill[blue!6,   rounded corners=3pt] (-1.9,-6.8) rectangle (15, 2.7);
      \draw[dashed, gray!60, line width=0.9pt] (6.7,-6.8) -- (6.7, 2.7);
      \draw[dashed, gray!60, line width=0.9pt] (-1.9,-2.3) -- (15,-2.3);
    \end{scope}

    \node[stagelbl] at ( 1.95, 2.2) {Stage 1: CPT{+}SFT};
    \node[stagelbl] at (11.0, 2.2) {Stage 2: Online M$^2$PO RL};
    \node[stagelbl] at (10.5,-3.0) {Stage 3: OPD};
    \node[stagelbl] at ( 1.95,-3.0) {Stage 4: Structure-Aware Inference and Postcheck};

    \node[data] (d1) at (-0.0, 1.4) {Multilingual\\(100+ langs)};
    \node[data] (d2) at (-0.0, 0.5) {WMT26\\per-direction};
    \node[data] (d3) at (-0.0,-0.4) {General\\instruction};
    \node[data] (d4) at (-0.0,-1.3) {Translation\\instruction};

    \node[bigbox, fill=blue!15, minimum width=15mm, minimum height=15mm]
      (sft) at (2.85, 0) {};
    \node[submod, fill=white, text width=10mm, minimum height=4mm]
      (cpt) at (2.85, 0.35) {CPT};
    \node[submod, fill=white, text width=10mm, minimum height=4mm]
      (sfti) at (2.85,-0.35) {SFT};
    \draw[ar] (cpt.south) -- (sfti.north);
    \node[font=\scriptsize\bfseries, above=0.8mm of sft] {Full parameter};

    \draw[thick, decorate, decoration={brace, amplitude=20pt, raise=2pt}]
      (d1.north east) -- (d4.south east);

    \node[model] (m14a) at (5.4, 0) {14B$_{\mathsf{CPT{+}SFT}}$};
    \draw[ar] (sft.east) -- (m14a.west);

    \node[submod, fill=orange!12, minimum height=8mm] (pol)  at (9.0, 1.1) {Policy $\pi_\theta$};
    \node[submod, fill=orange!12, minimum height=8mm] (sam)  at (12.5,1.1) {Sample\\$K{=}8$};
    \node[submod, fill=orange!12, minimum height=8mm] (rew)  at (12.5,-1.1)
      {DA reward\\(DeepSeek-V4-Pro)};
    \node[submod, fill=orange!12, minimum height=8mm] (loss) at (9.0, -1.1)
      {M$^2$PO loss\\(23 directions)};

    \draw[ar] (pol.east)   -- node[edgelbl] {$K$ cands} (sam.west);
    \draw[ar] (sam.south)  -- (rew.north);
    \draw[ar] (rew.west)   -- node[edgelbl] {scores}  (loss.east);
    \draw[ar] (loss.north) -- node[edgelbl] {update}  (pol.south);

    \draw[thick, rounded corners=8pt] (7.6,-1.75) rectangle (13.9,1.75);

    \draw[ar] (m14a.east) -- (7.6,0);

    \node[model] (m14b) at (9.0,-4.1) {14B$_{\mathsf{CPT{+}SFT{+}RL}}$};
    \draw[ar] (9.0,-1.75) -- (m14b.north);

    \node[submod, fill=violet!14, text width=19mm, minimum height=9mm]
      (tch) at (12.8,-4.1) {Teacher 32B\\(frozen)};
    \node[final, minimum height=9mm] (final) at (9.0,-6.1) {14B student};
    \draw[ardash] (tch.west)
      -- node[font=\tiny, above=0.2mm] {reverse KL} (m14b.east);
    \draw[ar] (m14b.south) -- (final.north);

    \node[submod, fill=teal!12, text width=20mm, minimum height=9mm]
      (chk) at (3.3,-6.1) {Rule-based checks\\\& repair};
    \node[submod, fill=teal!12, text width=20mm, minimum height=9mm]
      (rep) at (3.3,-4.3) {Model rewrite\\(Qwen3.5-4B)};
    \node[final, fill=teal!35, text width=19mm, minimum height=9mm]
      (out) at (0.2,-5.2) {Translation\\output};

    \draw[ar] (final.west)
      -- node[font=\tiny, pos=0.66, above=0.3mm, align=center]
         {structure-aware decode} (chk.east);
    \draw[ardash] (chk.north) -- node[edgelbl] {fail} (rep.south);
    \draw[ar] (chk.west) -- node[font=\tiny, above=-0.6mm] {pass}
      (0.2,-6.1) -- (out.south);
    \draw[ar] (rep.west) -- (0.2,-4.3) -- (out.north);

  \end{tikzpicture}%
  }
  \caption{Overall pipeline of Lumen. \textbf{Stage~1 (CPT{+}SFT)}
  adapts a Qwen3-14B backbone to multilingual translation and
  translation-specific instruction following. \textbf{Stage~2 (online
  M\textsuperscript{2}PO)} uses direct-assessment rewards from
  DeepSeek-V4-Pro to optimize the current policy from sampled candidates.
  \textbf{Stage~3 (OPD)} distills instruction-following behavior from a
  frozen Qwen3-32B-Instruct teacher into the 14B student. \textbf{Stage~4}
  performs structure-aware decoding and conservative postchecking to
  preserve required formats and repair clear failures before submission.}
  \label{fig:pipeline}
\end{figure*}

Lumen comprises four successive stages, as illustrated in
Figure~\ref{fig:pipeline}. The first stage adapts the Qwen3-14B
foundation model through continued pre-training and supervised
fine-tuning, jointly strengthening multilingual translation and
translation-specific instruction following. The second stage extends
M\textsuperscript{2}PO to online preference optimization, using
on-policy candidates and LLM-based direct-assessment rewards to improve
translation fidelity. The third stage applies on-policy distillation to
transfer instruction-following behavior from a larger frozen teacher to
the translation-specialized 14B model. The final stage performs
structure-aware decoding and targeted postchecking to preserve required
formats and repair clear decoding failures.

\subsection{CPT and SFT Adaptation}
\label{sec:sft}

In Stage~1, we adapt the Qwen3-14B foundation model in two consecutive
full-parameter phases (Figure~\ref{fig:pipeline}, Stage~1). The first
phase is continued pre-training (CPT) on multilingual text covering the
task-relevant domains and 23 translation directions, supplemented with
general instruction data. This phase strengthens domain and language
coverage, particularly for low-resource and regional languages that are
underrepresented in general pre-training corpora. The second phase is
supervised fine-tuning (SFT) on a mixture of parallel translation and
instruction-bearing data. Both phases update all model parameters: CPT
strengthens the multilingual translation foundation, while SFT jointly
develops translation quality and translation-specific instruction
following.

\subsubsection{Training Data}
\paragraph{CPT data.} The continued-pre-training corpus combines
multilingual text from task-relevant domains and all 23 translation
directions with general instruction data. The multilingual component is
constructed from monolingual text and the source sides of parallel
corpora drawn from internal resources, WMT releases, and public
multilingual resources such as FLORES+ \citep{nllb2024}. We cap and
stratify high-resource corpora to prevent them from dominating the
mixture, while increasing the representation of low-resource and
regional languages.

\paragraph{SFT data.} The SFT mixture contains four sources:
multilingual parallel data, task-specific parallel data for the 23
WMT26 directions, general instruction data, and translation-specific
instruction data. Parallel translation remains the majority of the
mixture to preserve the model's translation competence, while the two
instruction sources provide general and translation-specific
instruction-following supervision.
When reliable references are unavailable, we supplement the parallel
data with model-generated translations retained after quality
filtering, with additional synthetic data used to improve the coverage
of low-resource directions.

All corpora undergo normalization, deduplication, language and script
verification, and length-based filtering. We additionally score
parallel pairs with XCOMET-XXL \citep{guerreiro2024xcomet}, from the COMET
family of learned MT metrics \citep{rei2020comet}, and remove
low-confidence translations before training.

\paragraph{Translation-specific instruction data.}
Translation-specific instruction data is the core of our
instruction-following adaptation. Each example augments a source--target
pair with natural-language constraints specifying the requirements to be
satisfied. Table~\ref{tab:instr-types} summarizes the training taxonomy.

\begin{table*}[t]
  \centering
  \small
  \begin{tabular}{@{}p{0.30\textwidth}p{0.64\textwidth}@{}}
    \toprule
    \textbf{Category} & \textbf{Definition} \\
    \midrule
    Literary (novel) &
    Preserve narrative voice, dialogue, and cultural nuance. \\
    \addlinespace
    Game &
    Translate game dialogue and UI text under output constraints. \\
    \addlinespace
    Medical &
    Preserve medical terminology, drug names, codes, and entities. \\
    \addlinespace
    General-domain instruction MT &
    Follow content and formatting requirements in general-domain MT. \\
    \addlinespace
    Structured output &
    Preserve tags, identifiers, and parseable output structure. \\
    \addlinespace
    Style &
    Produce the translation in a specified tone or register. \\
    \addlinespace
    General instruction &
    Retain broad multilingual instruction-following ability. \\
    \bottomrule
  \end{tabular}
  \caption{Instruction-bearing data used in SFT.}
  \label{tab:instr-types}
\end{table*}

Instruction-bearing examples constitute a minority of the SFT mixture,
whose main component remains parallel translation data. The taxonomy
covers the four WMT26 instruction types: formal or informal voice,
glossary constraints, structured translation, and style and expression
requirements. These four types provide the core translation-instruction
supervision; the domain-specific and general-instruction categories in
Table~\ref{tab:instr-types} serve as auxiliary data that broaden domain
coverage and retain general instruction-following ability.

Because manual construction is costly, most translation-specific
instruction examples are produced through an LLM-assisted pipeline. We
sample high-quality parallel pairs, use Qwen3.7-Max to generate diverse
constraints and adapt the corresponding examples, and use the same model
in a separate judging pass to filter non-compliant samples. Diversity is
controlled across instruction templates, domains, and style intensity.

\subsubsection{Adaptation Parameterization}
Both CPT and SFT update all model parameters rather than using
parameter-efficient adapters. For the SFT phase, we use a learning rate
of $2\times10^{-5}$, a global batch size of $1{,}152$ sequences, and the
AdamW optimizer ($\beta_1=0.9$, $\beta_2=0.999$). The resulting CPT+SFT
model is used by the subsequent RL and OPD stages.

\subsection{Preference Alignment with Online M\textsuperscript{2}PO}
\label{sec:rl}

M\textsuperscript{2}PO (Multi-Perspective Multi-Pair Preference
Optimization) \citep{m2po2026} was originally formulated as an
\emph{offline} preference-optimization method over a fixed set of ranked
candidates. We extend it to an online setting: at each update, the
current policy generates fresh candidates, an LLM judge ranks them, and
the resulting preferences are immediately used to update the policy.
This retains M\textsuperscript{2}PO's emphasis on translation fidelity
and its multi-pair supervision, while avoiding reliance on a static
preference dataset.

\subsubsection{Online M\textsuperscript{2}PO}
\label{sec:rl-m2po}
For an input $x$, the current policy $\pi_\theta$ samples $K=8$
candidates $\{y_1,\ldots,y_8\}$. Each candidate receives a
direct-assessment (DA) score from an LLM judge and is ranked by this
score. Let $y_{(1)},\ldots,y_{(K)}$ denote the candidates in descending
score order. M\textsuperscript{2}PO constructs $K/2$ non-overlapping
preference pairs by coupling the two ends of the ranked list, with
\(\mathcal{P}(x)=\{(i,K-i+1):1\leq i\leq K/2\}\). Thus, $K=8$ yields
four preference pairs. The M\textsuperscript{2}PO loss is
\begin{equation}
  \mathcal{L}_{\text{M}^2\text{PO}} = -\mathbb{E}_{x}
  \sum_{(i,j)\in\mathcal{P}(x)} w_{ij}\,
  \log \sigma\!\big(\beta\delta_{ij}-m_{ij}\big),
  \label{eq:m2po}
\end{equation}
where $\delta_{ij} = \Delta_{ij}\log\pi_\theta-
\Delta_{ij}\log\pi_{\text{ref}}$, with
$\Delta_{ij}\log\pi=\log\pi(y_{(i)}\mid x)-
\log\pi(y_{(j)}\mid x)$. Here, $m_{ij}$ is a margin proportional to the
reward difference, $w_{ij}$ is a pair weighting coefficient, and
$\pi_{\text{ref}}$ is the frozen
CPT+SFT model. Because candidates are drawn from the current policy at
every step, the preference data evolves with the model rather than being
fixed in advance.

\paragraph{LLM-judge reward.} We use DeepSeek-V4-Pro\footnote{\url{https://huggingface.co/deepseek-ai/DeepSeek-V4-Pro}} as a
direct-assessment judge \citep{zheng2023judging,kocmi2023gemba}. For each
candidate, the judge returns a single integer between $0$ and $100$ that
assesses meaning preservation and grammatical quality. When a reference
translation is available, it is included as auxiliary evidence;
otherwise, the candidate is assessed against the source alone. We
normalize the score as
\begin{equation}
  r(x, y) = \frac{\text{DA}(x, y)}{100} \in [0, 1],
  \label{eq:reward}
\end{equation}
where $\text{DA}(x,y)$ denotes the judge score. The continuous reward
preserves the ordering among intermediate-quality candidates and sets the
pair margins in Eq.~\ref{eq:m2po}.

\subsubsection{Training Configuration}
\label{sec:rl-config}
We optimize the CPT+SFT model with LoRA adapters~\citep{hu2022lora} (rank
$128$, $\alpha=256$) and merge the final adapter back into the full model
after training. We use a learning rate of $5\times10^{-5}$ and the AdamW
optimizer ($\beta_1=0.9$, $\beta_2=0.999$). Each optimization step
processes $8$ prompt groups ($64$ sequences in total); each group
consists of one prompt and its $K=8$ candidates, sampled at temperature
$1.0$ using vLLM~\citep{kwon2023vllm} for efficient high-throughput
generation. The preference temperature in Eq.~\ref{eq:m2po} is set to
$\beta=0.1$.

\subsection{On-Policy Distillation for Translation Instruction Following}
\label{sec:opd}

After CPT, SFT, and online M\textsuperscript{2}PO, our Qwen3-14B reaches
strong translation quality, but residual errors remain in following
fine-grained constraints on style, terminology, and output format. This
gap is expected because the DA reward used in the online
M\textsuperscript{2}PO stage assesses translation quality rather than
instruction compliance. We therefore adapt on-policy distillation (OPD)
to translation-instruction data. The frozen teacher $\pi_T$ is the
off-the-shelf open-source Qwen3-32B-Instruct model\footnote{\url{https://huggingface.co/Qwen/Qwen3-32B}};
the student $\pi_\theta$ is our Qwen3-14B model after CPT, SFT, and
online M\textsuperscript{2}PO. We keep the teacher frozen and use it
only as a source of instruction-following supervision. Thus, OPD can
transfer behavior from the larger model while preserving the
translation-specialized 14B student as the final inference model.

\subsubsection{Distillation Objective}
\label{sec:opd-method}
OPD differs from conventional sequence-level distillation over a fixed
teacher-generated dataset \citep{hinton2015distilling}. For each
translation-instruction input $x$, the current student first samples a
translation $y \sim \pi_\theta(\cdot\mid x)$. The teacher then evaluates
the same trajectory and supplies token-level distributions at each
position. Let $p_t$ and $q_t$ denote the student and teacher next-token
distributions conditioned on $(x,y_{<t})$, respectively. We minimize the
reverse KL along the student trajectory:
\begin{equation}
  \mathcal{L}_{\text{OPD}} = \mathbb{E}_{x}\,
  \mathbb{E}_{y \sim \pi_\theta(\cdot \mid x)}
  \!\left[ \frac{1}{|y|} \sum_{t=1}^{|y|}
  \mathrm{KL}\big( p_t \,\|\, q_t \big) \right]\!,
  \label{eq:opd}
\end{equation}
Because training trajectories are generated by the current student, the
teacher supervises the distribution where the student actually makes
errors, rather than only a fixed set of teacher outputs. We use reverse
KL, which encourages the student to place probability mass on the
teacher's preferred continuations. The two models share the Qwen3
tokenizer and vocabulary, so the token-level KL is computed directly
without vocabulary alignment.

\subsubsection{Training Configuration}
\label{sec:opd-config}
We fine-tune the full student model with this reverse-KL objective,
following on-policy distillation \citep{gu2024minillm}. We use AdamW
($\beta_1=0.9$, $\beta_2=0.999$), a learning rate of $1\times10^{-5}$,
and a global batch size of $256$. The learning rate is lower than that
used for SFT because OPD refines the student's output distribution rather
than teaching translation from scratch. Student trajectories are sampled
at temperature $1.0$, and both teacher and student use a distillation
temperature of $1.0$. For each trajectory, the frozen teacher supplies
its token-level distribution in a single forward pass.
\begin{table*}[t]
  \centering
  \small
  \setlength{\tabcolsep}{6pt}
  \begin{tabular}{@{}lrrrrrr@{}}
    \toprule
    & \multicolumn{2}{c}{\textbf{Instr.\ following}}
    & \multicolumn{2}{c}{\textbf{Quality}}
    & \multicolumn{2}{c}{\textbf{Overall}} \\
    \cmidrule(lr){2-3}\cmidrule(lr){4-5}\cmidrule(lr){6-7}
    \textbf{System} & \textbf{Avg} & \textbf{Pass\%} &
    \textbf{Avg} & \textbf{Pass\%} & \textbf{Avg} & \textbf{Pass\%} \\
    \midrule
    Qwen3.7-Max-Think    & 4.50 & 83.8 & 4.76 & 94.0 & 4.63 & 88.9 \\
    GPT-5.5              & 4.49 & 83.4 & 4.77 & 94.0 & 4.63 & 88.7 \\
    Gemini-3.1-Pro       & 4.49 & 83.7 & 4.70 & 91.1 & 4.59 & 87.4 \\
    Qwen3.7-Max-NoThink  & 4.32 & 78.4 & 4.48 & 85.3 & 4.40 & 81.8 \\
    \midrule
    \textbf{Ours (14B)}  & 4.49 & 83.4 & 4.71 & 92.0 & 4.60 & 87.7 \\
    \bottomrule
  \end{tabular}
  \caption{Main results on the WMT26 test set (23 official directions).
  For each aspect we report the mean $1$--$5$ rubric score (Avg) and the
  pass rate (Pass\%, percentage of outputs scoring $\geq 4$), both judged
  by DeepSeek-V4-Pro (Section~\ref{sec:exp-eval}); Overall Avg and Pass\% are
  the unweighted averages of the corresponding values across the two
  aspects. Baseline rows are ordered by Overall Avg; our system is
  listed separately in the last row.}
  \label{tab:main}
\end{table*}

\subsection{Structure-Aware Inference and Postcheck}
\label{sec:postcheck}

Stage~4 is applied after the distilled 14B model produces a translation.
It has two goals: preserve non-linguistic structure required by the input
and prevent clear decoding failures from entering the final submission.
We decode with greedy search using vLLM, then apply structure-aware
processing and a conservative postcheck. This stage is deliberately
limited to objective failures; it is not intended as a general-purpose
post-editor.

\paragraph{Structure-aware decoding.} We first identify whether an input
is plain text, HTML, or JSON. For HTML, we translate text nodes and
designated translatable attributes; for JSON, we translate string leaves.
The translated fragments are then reassembled into the original object,
preserving markup, keys, and protected content such as URLs, e-mail
addresses, and numbers. Long plain-text inputs are split at
sentence-aware boundaries, translated chunk by chunk, and concatenated.

\paragraph{Postcheck and repair.} Each output is checked for abnormal
source--target length ratios, structural mismatches (HTML tag signatures
or JSON shape), repeated character or $n$-gram loops, insufficient
target-script coverage, and unintended copying of the source. A failed
check does not automatically imply a translation error; it identifies a
small set of hypotheses for review. When a flagged defect is mechanical
and has an unambiguous fix, such as a missing code-fence marker, we apply
a deterministic repair. The remaining flagged outputs are regenerated
with Qwen3.5-4B-Instruct\footnote{\url{https://huggingface.co/Qwen/Qwen3.5-4B}}
under the original input and structural constraints,
whereas outputs that pass all checks are retained unchanged. This routing
keeps repair focused on degenerate or structurally invalid outputs and
avoids unnecessary changes to otherwise adequate translations.

\section{Experiments}
\label{sec:experiments}

\begin{table}[t]
  \centering
  \footnotesize
  \setlength{\tabcolsep}{4pt}
  \begin{tabular}{@{}lrrrr@{}}
    \toprule
    & \multicolumn{2}{c}{\textbf{Instr.\ follow.}}
    & \multicolumn{2}{c}{\textbf{Quality}} \\
    \cmidrule(lr){2-3}\cmidrule(lr){4-5}
    \textbf{Dir.} & Avg & Pass\% & Avg & Pass\% \\
    \midrule
    cs$\rightarrow$de                   & 4.90 & 96.5 & 4.56 & 86.7 \\
    cs$\rightarrow$uk                   & 4.94 & 98.7 & 4.55 & 86.3 \\
    cs$\rightarrow$vi                   & 4.94 & 97.5 & 4.43 & 80.3 \\
    zh$\rightarrow$ja                   & 4.96 & 98.4 & 4.40 & 77.7 \\
    en$\rightarrow$ar\textsubscript{EG} & 4.92 & 97.1 & 4.66 & 89.7 \\
    en$\rightarrow$zh\textsubscript{CN} & 4.97 & 99.7 & 4.88 & 96.2 \\
    en$\rightarrow$zh\textsubscript{TW} & 4.97 & 98.5 & 4.84 & 95.5 \\
    en$\rightarrow$cs                   & 3.99 & 64.0 & 4.73 & 94.4 \\
    en$\rightarrow$et                   & 3.77 & 68.4 & 4.65 & 90.8 \\
    en$\rightarrow$de                   & 4.29 & 81.9 & 4.92 & 97.7 \\
    en$\rightarrow$is                   & 4.98 & 99.4 & 4.79 & 93.6 \\
    en$\rightarrow$ja                   & 4.99 &100.0 & 4.71 & 89.9 \\
    en$\rightarrow$ko                   & 4.93 & 97.6 & 4.82 & 94.2 \\
    en$\rightarrow$ru                   & 3.97 & 60.6 & 4.60 & 91.0 \\
    en$\rightarrow$uk                   & 4.93 & 98.5 & 4.69 & 93.9 \\
    en$\rightarrow$hy                   & 4.96 & 99.0 & 4.63 & 88.4 \\
    en$\rightarrow$be                   & 4.94 & 99.5 & 4.68 & 90.4 \\
    en$\rightarrow$id                   & 4.96 & 99.5 & 4.80 & 94.9 \\
    en$\rightarrow$kk                   & 4.98 & 99.5 & 4.62 & 88.9 \\
    en$\rightarrow$th                   & 4.97 & 99.5 & 4.79 & 93.4 \\
    en$\rightarrow$lij                  & 4.97 & 99.5 & 4.44 & 76.8 \\
    en$\rightarrow$se                   & 4.97 & 99.5 & 4.12 & 71.7 \\
    en$\rightarrow$lld                  & 4.99 &100.0 & 4.24 & 69.7 \\
    \midrule
    \textbf{Overall}                    & \textbf{4.49} & \textbf{83.4} & \textbf{4.71} & \textbf{92.0} \\
    \bottomrule
  \end{tabular}
  \caption{Per-direction results of our submitted 14B system on the 23 official WMT26 directions. For each aspect we
  report the mean $1$--$5$ rubric score (Avg) and the pass rate (Pass\%,
  percentage of outputs scoring $\geq 4$), judged by DeepSeek-V4-Pro. Overall is the
  micro-average over all official test items.}
  \label{tab:perdir}
\end{table}

\begin{table}[t]
  \centering
  \small
  \setlength{\tabcolsep}{4pt}
  \begin{tabular}{@{}lrrrr@{}}
    \toprule
    & \multicolumn{2}{c}{\textbf{Instr.\ following}}
    & \multicolumn{2}{c}{\textbf{Quality}} \\
    \cmidrule(lr){2-3}\cmidrule(lr){4-5}
    \textbf{Configuration} & \textbf{Avg} & \textbf{Pass\%} &
    \textbf{Avg} & \textbf{Pass\%} \\
    \midrule
    Base (14B) & 1.14 & 0.6 & 1.09 & 1.0 \\
    + CPT+SFT & 4.08 & 69.5 & 4.23 & 74.8 \\
    + Online M$^2$PO & 4.13 & 76.8 & 4.55 & 89.4 \\
    + OPD & 4.38 & 82.1 & 4.67 & 91.3 \\
    + Postcheck & \textbf{4.49} & \textbf{83.4} & \textbf{4.71} & \textbf{92.0} \\
    \bottomrule
  \end{tabular}
  \caption{Cumulative ablation on the WMT26 test set. For each aspect we
  report the mean $1$--$5$ rubric score (Avg) and the pass rate (Pass\%,
  percentage of outputs scoring $\geq 4$), both judged by DeepSeek-V4-Pro
  (Section~\ref{sec:exp-eval}).}
  \label{tab:ablation}
\end{table}

\subsection{Experimental Setup and Evaluation}
\label{sec:exp-setup}

\paragraph{Systems and Test Sets.}
Our submission combines the distilled
Qwen3-14B-Base\footnote{\url{https://huggingface.co/Qwen/Qwen3-14B-Base}}
(CPT+SFT+RL+OPD) with the structure-aware inference and postcheck stage
described in Section~\ref{sec:system}. We evaluate on the official WMT26
General MT test sets across the 23 participated directions.

We compare against GPT-5.5, Qwen3.7-Max (thinking mode), Gemini-3.1-Pro,
and Qwen3.7-Max (non-thinking mode), all evaluated
under the same protocol. We also report internal ablations
(Section~\ref{sec:exp-abl}).

\paragraph{Evaluation Protocol.}
\label{sec:exp-eval}
Because official WMT26 human-evaluation results were not available at
submission time, we use an automatic \emph{LLM-as-judge} protocol. Each
output is scored independently on an absolute $1$--$5$ rubric along two
aspects: instruction following and translation quality. We report the
mean score and the pass rate, defined as the proportion of outputs with a
score of at least $4$.

For instruction following, the judge receives only the task instruction
and system output, and evaluates target language, output-only compliance,
glossary use, register, and required structure such as
JSON, HTML, tags, or placeholders. For translation quality, it evaluates
semantic accuracy and completeness, terminology, and fluency. A reference
translation, when available, is supplied only as auxiliary evidence; the
judge is instructed to assess the output against the source rather than
penalize legitimate differences from the reference.

Both aspects use DeepSeek-V4-Pro at temperature $0$. For each item, the judge
returns a brief justification and a score in the fixed form
\texttt{[[score=X]]}, where $X\in\{1,\dots,5\}$. The
instruction-following evaluation set covers all 23 directions and the
four official WMT26 instruction categories: formal/informal voice,
glossaries, structured translation, and style and expressions.

\subsection{Main Results}
\label{sec:exp-main}
Table~\ref{tab:main} reports every system's instruction-following and
translation-quality scores under the absolute $1$--$5$ rubric of
Section~\ref{sec:exp-eval}. Among the four reference LLMs, Qwen3.7-Max
(thinking) and GPT-5.5 lead with essentially tied overall means
($4.63$ and $4.63$), followed by Gemini-3.1-Pro ($4.59$);
Qwen3.7-Max in non-thinking mode
trails clearly ($4.40$). Quality scores are
uniformly high ($4.48$--$4.77$), so the systems separate mainly on the
harder instruction-following aspect. Our distilled 14B system attains an
overall Avg of $4.60$ and Overall Pass of $87.7\%$. Despite its 14B
translation backbone, it is comparable to the leading proprietary systems
on translation-instruction following (Avg $4.49$, Pass $83.4\%$, versus
$4.49$/$83.4\%$ for GPT-5.5 and $4.50$/$83.8\%$ for
Qwen3.7-Max-Think), while its translation quality ($4.71$, Pass $92.0\%$)
surpasses Gemini-3.1-Pro.

\subsection{Per-Direction Results}
\label{sec:exp-perdir}
Table~\ref{tab:perdir} breaks down our submitted system's scores across
the 23 official directions. Instruction following is near-ceiling on the
large majority of directions (mean $\geq\!4.9$ with pass rates above
$97\%$), indicating that the instruction-specialized pipeline transfers
robustly across languages and scripts. The clear exceptions are
en$\rightarrow$et ($3.77$), en$\rightarrow$ru ($3.97$),
en$\rightarrow$cs ($3.99$), and to a lesser extent en$\rightarrow$de
($4.29$). These directions also correspond to the largest test sets,
with more than one thousand items each, and therefore contribute
substantially to the aggregate instruction-following score. Translation quality is uniformly high
(mostly $\geq\!4.6$) and dips only on the lowest-resource new
directions---en$\rightarrow$se ($4.12$), en$\rightarrow$lld ($4.24$),
zh$\rightarrow$ja ($4.40$), cs$\rightarrow$vi ($4.43$), and
en$\rightarrow$lij ($4.44$). The two weak spots are largely disjoint:
directions with lower instruction-following scores are generally not
those with the lowest translation-quality scores. This separation
suggests that the two evaluation aspects capture different failure modes.

\subsection{Ablation Study}
\label{sec:exp-abl}
We conduct a cumulative ablation to quantify the incremental effects of
the training stages and final postcheck (Table~\ref{tab:ablation}). Both
translation-instruction following and translation quality improve
monotonically across the cumulative configurations, although the largest
gain of each stage occurs on a different dimension.

\paragraph{CPT+SFT establishes the task capability.} The unadapted
Qwen3-14B model performs poorly under the translation-instruction
prompts (instruction following: $1.14$ Avg and $0.6\%$ Pass; translation
quality: $1.09$ Avg and $1.0\%$ Pass). Adding CPT+SFT raises instruction
following to $4.08$ Avg and $69.5\%$ Pass, and translation quality to
$4.23$ Avg and $74.8\%$ Pass. Relative to the base model, this amounts to
gains of $2.94$ and $3.14$ rubric points, respectively. This low starting
point is expected: the base is a pre-trained Qwen3-14B model without
task-specific translation adaptation or translation-instruction
supervision. The result shows that WMT-aligned continued pre-training and
instruction-bearing SFT data are necessary to turn this general backbone
into a usable instruction-specialized translation system.

\paragraph{Online M\textsuperscript{2}PO primarily improves translation
quality.} On top of CPT+SFT, online preference optimization raises
translation quality from $4.23$ to $4.55$ Avg and from $74.8\%$ to
$89.4\%$ Pass. The corresponding instruction-following change is more
modest ($4.08\!\rightarrow\!4.13$ Avg; $69.5\%\!\rightarrow\!76.8\%$
Pass). This asymmetric effect aligns with the direct-assessment reward,
which ranks candidates by translation quality and provides the most
direct supervision for adequacy, completeness, and fluency. The online
extension also continually refreshes the candidate set from the current
policy, allowing the M\textsuperscript{2}PO objective to target errors
that remain after CPT+SFT.

\paragraph{OPD closes the remaining instruction-following gap.} Starting
from the CPT+SFT+online M\textsuperscript{2}PO model, OPD increases
instruction following from $4.13$ to $4.38$ Avg and from $76.8\%$ to
$82.1\%$ Pass, while translation quality continues to improve from $4.55$
to $4.67$ Avg and from $89.4\%$ to $91.3\%$ Pass. Thus, the teacher's
instruction-following behavior transfers without sacrificing the quality
improvements obtained in the preceding stage. The larger teacher is used
only during training; translation at inference is performed by the 14B
student.

\paragraph{Postcheck recovers a small set of severe failures.} The
postcheck is applied after all training stages. It further increases
instruction following from $4.38$ to $4.49$ Avg and from $82.1\%$ to
$83.4\%$ Pass, while translation quality rises from $4.67$ to $4.71$ Avg
and from $91.3\%$ to $92.0\%$ Pass. On \texttt{en-zh\_CN}, it altered
$31$ of $945$ hypotheses ($3.3\%$): $7$ received deterministic fence
restoration, and $24$ were selectively rewritten across two review
rounds. These outputs were identified by objective structural or decoding
signals, such as malformed structure, missing lines, or degenerate
generation. Thus, the postcheck obtains its gains by recovering a small
number of high-impact failures rather than by broadly rewriting otherwise
adequate translations.

\subsection{Analysis}
\label{sec:exp-analysis}
The results reveal a clear division of labor across the training stages.
CPT+SFT establishes the basic ability to translate under explicit
instructions, online M\textsuperscript{2}PO supplies the largest
subsequent improvement in translation quality, and OPD contributes the
largest later-stage gain in instruction following. The complementary
pattern is important: optimizing only direct-assessment translation
quality is insufficient to reach the final instruction-following level,
whereas distillation improves instruction adherence without reversing the
quality gains obtained by online preference optimization. The final
postcheck contributes a further inference-time improvement by recovering
the small tail of structurally invalid or degenerate outputs that remain
after decoding.

The per-direction results further show that instruction following and
translation quality do not fail on exactly the same directions. This
supports evaluating them separately rather than collapsing them into a
single automatic score. At the system level, Lumen's instruction-following
score is close to GPT-5.5 and Qwen3.7-Max-Think, while its translation
quality remains within the range of the compared general-purpose systems.
Taken together, these results support the design of a translation-
specialized 14B system that combines quality-oriented preference
optimization with a dedicated mechanism for instruction following.

\section{Conclusion}
\label{sec:conclusion}
We presented Lumen, our submission to the WMT2026 General
Machine Translation Shared Task. Targeting instruction-specialized
translation, we introduced a four-stage pipeline: instruction-specialized
CPT+SFT, where continued pre-training on a corpus following the WMT
domains and language directions (plus general instruction data) is
followed by supervised fine-tuning that jointly builds translation and
instruction-following ability;
online preference optimization with M\textsuperscript{2}PO and a
direct-assessment LLM-judge reward; on-policy distillation that transfers
the instruction-following ability of an off-the-shelf Qwen3-32B-Instruct
teacher into our efficient 14B student; and structure-aware inference with
targeted postchecking and repair for clear structural or decoding failures.
Under an absolute
$1$--$5$ LLM-judge rubric that scores instruction following and translation
quality separately, our 14B system reaches an overall mean of $4.60$.
Despite using a 14B translation backbone, Lumen achieves performance
comparable to leading proprietary systems in both translation quality and
translation-instruction following, showing that the proposed pipeline can
narrow the gap to substantially larger closed systems.

\bibliography{custom}

\begin{thebibliography}{20}
\providecommand{\natexlab}[1]{#1}

\bibitem[{Alves et~al.(2024)Alves, Pombal, Guerreiro, Martins
  et~al.}]{alves2024tower}
Duarte~M. Alves, Jos{\'e} Pombal, Nuno~M. Guerreiro, Andr{\'e} F.~T. Martins,
  and 1 others. 2024.
\newblock Tower: An open multilingual large language model for
  translation-related tasks.
\newblock In \emph{Conference on Language Modeling (COLM)}.

\bibitem[{Gu et~al.(2024)Gu, Dong, Wei, and Huang}]{gu2024minillm}
Yuxian Gu, Li~Dong, Furu Wei, and Minlie Huang. 2024.
\newblock {MiniLLM}: Knowledge distillation of large language models.
\newblock In \emph{International Conference on Learning Representations
  (ICLR)}.

\bibitem[{Guerreiro et~al.(2024)Guerreiro, Rei, van Stigt, Coheur, Colombo, and
  Martins}]{guerreiro2024xcomet}
Nuno~M. Guerreiro, Ricardo Rei, Daan van Stigt, Luisa Coheur, Pierre Colombo,
  and Andr{\'e} F.~T. Martins. 2024.
\newblock {xCOMET}: Transparent machine translation evaluation through
  fine-grained error detection.
\newblock \emph{Transactions of the Association for Computational Linguistics
  (TACL)}, 12:979--995.

\bibitem[{Hendy et~al.(2023)Hendy, Abdelrehim, Sharaf, Raunak, Gabr,
  Matsushita, Kim, Afify, and Awadalla}]{hendy2023how}
Amr Hendy, Mohamed Abdelrehim, Amr Sharaf, Vikas Raunak, Mohamed Gabr, Hitokazu
  Matsushita, Young~Jin Kim, Mohamed Afify, and Hany~Hassan Awadalla. 2023.
\newblock How good are {GPT} models at machine translation? a comprehensive
  evaluation.
\newblock \emph{arXiv preprint arXiv:2302.09210}.

\bibitem[{Hinton et~al.(2015)Hinton, Vinyals, and Dean}]{hinton2015distilling}
Geoffrey Hinton, Oriol Vinyals, and Jeff Dean. 2015.
\newblock Distilling the knowledge in a neural network.
\newblock \emph{arXiv preprint arXiv:1503.02531}.

\bibitem[{Hu et~al.(2022)Hu, Shen, Wallis, Allen-Zhu, Li, Wang, Wang, and
  Chen}]{hu2022lora}
Edward~J. Hu, Yelong Shen, Phillip Wallis, Zeyuan Allen-Zhu, Yuanzhi Li, Shean
  Wang, Lu~Wang, and Weizhu Chen. 2022.
\newblock {LoRA}: Low-rank adaptation of large language models.
\newblock In \emph{International Conference on Learning Representations
  (ICLR)}.

\bibitem[{Kim and Rush(2016)}]{kim2016sequence}
Yoon Kim and Alexander~M. Rush. 2016.
\newblock Sequence-level knowledge distillation.
\newblock In \emph{Proceedings of the 2016 Conference on Empirical Methods in
  Natural Language Processing (EMNLP)}.

\bibitem[{Kocmi et~al.(2024)Kocmi, Avramidis, Bawden, Bojar, Dvorkovich,
  Federmann, Fishel, Freitag, Gowda, Grundkiewicz et~al.}]{kocmi2024wmt24}
Tom Kocmi, Eleftherios Avramidis, Rachel Bawden, Ond{\v{r}}ej Bojar, Anton
  Dvorkovich, Christian Federmann, Mark Fishel, Markus Freitag, Thamme Gowda,
  Roman Grundkiewicz, and 1 others. 2024.
\newblock Findings of the wmt24 general machine translation shared task: The
  llm era is here but mt is not solved yet.
\newblock In \emph{Proceedings of the Ninth Conference on Machine Translation},
  pages 1--46.

\bibitem[{Kocmi and Federmann(2023)}]{kocmi2023gemba}
Tom Kocmi and Christian Federmann. 2023.
\newblock Large language models are state-of-the-art evaluators of translation
  quality.
\newblock In \emph{Proceedings of the 24th Annual Conference of the European
  Association for Machine Translation (EAMT)}.

\bibitem[{Kocmi et~al.(2025)}]{kocmi2025wmt25}
Tom Kocmi and 1 others. 2025.
\newblock \href {https://aclanthology.org/2025.wmt-1.22/} {Findings of the
  {WMT}25 general machine translation shared task: Time to stop evaluating on
  easy test sets}.
\newblock In \emph{Proceedings of the Tenth Conference on Machine Translation},
  Suzhou, China. Association for Computational Linguistics.

\bibitem[{Kwon et~al.(2023)Kwon, Li, Zhuang, Sheng, Zheng, Yu, Gonzalez, Zhang,
  and Stoica}]{kwon2023vllm}
Woosuk Kwon, Zhuohan Li, Siyuan Zhuang, Ying Sheng, Lianmin Zheng, Cody~Hao Yu,
  Joseph~E. Gonzalez, Hao Zhang, and Ion Stoica. 2023.
\newblock Efficient memory management for large language model serving with
  {PagedAttention}.
\newblock In \emph{Proceedings of the 29th Symposium on Operating Systems
  Principles (SOSP)}.

\bibitem[{{NLLB Team}(2024)}]{nllb2024}
{NLLB Team}. 2024.
\newblock Scaling neural machine translation to 200 languages.
\newblock \emph{Nature}, 630(8018):841--846.

\bibitem[{Rei et~al.(2020)Rei, Stewart, Farinha, and Lavie}]{rei2020comet}
Ricardo Rei, Craig Stewart, Ana~C Farinha, and Alon Lavie. 2020.
\newblock {COMET}: A neural framework for {MT} evaluation.
\newblock In \emph{Proceedings of the 2020 Conference on Empirical Methods in
  Natural Language Processing (EMNLP)}.

\bibitem[{Vaswani et~al.(2017)Vaswani, Shazeer, Parmar, Uszkoreit, Jones,
  Gomez, Kaiser, and Polosukhin}]{vaswani2017attention}
Ashish Vaswani, Noam Shazeer, Niki Parmar, Jakob Uszkoreit, Llion Jones,
  Aidan~N. Gomez, {\L}ukasz Kaiser, and Illia Polosukhin. 2017.
\newblock Attention is all you need.
\newblock In \emph{Advances in Neural Information Processing Systems
  (NeurIPS)}.

\bibitem[{Wang et~al.(2026)Wang, Xu, Liu, Liu, Zhao, Zeng, Shao, Dong, Wu,
  Zhou, Dong, Zeng, Wang, and Luo}]{m2po2026}
Hao Wang, Linlong Xu, Heng Liu, Yangyang Liu, Xiaohu Zhao, Bo~Zeng, Liangying
  Shao, Yichen Dong, Xinwei Wu, Jiang Zhou, Tianyu Dong, Xiangxiang Zeng,
  Longyue Wang, and Weihua Luo. 2026.
\newblock \href {https://doi.org/10.18653/v1/2026.acl-long.469} {{M}$^2${PO}:
  Multi-perspective multi-pair preference optimization for machine
  translation}.
\newblock In \emph{Proceedings of the 64th Annual Meeting of the Association
  for Computational Linguistics (Volume 1: Long Papers)}, pages 10315--10336,
  San Diego, California, United States. Association for Computational
  Linguistics.

\bibitem[{Wang et~al.(2025)Wang, Xu, Liu, Liu, Zhao, Zeng, Wang, Luo, and
  Zhang}]{wang2025marco}
Hao Wang, Linlong Xu, Heng Liu, Yangyang Liu, Xiaohu Zhao, Bo~Zeng, Longyue
  Wang, Weihua Luo, and Kaifu Zhang. 2025.
\newblock \href {https://aclanthology.org/2025.wmt-1.33/} {Marco large
  translation model at {WMT}2025: Transforming translation capability in {LLM}s
  via quality-aware training and decoding}.
\newblock In \emph{Proceedings of the Tenth Conference on Machine Translation},
  pages 587--593, Suzhou, China. Association for Computational Linguistics.

\bibitem[{Yang et~al.(2025)Yang, Li, Yang, Zhang, Hui, Zheng, Yu, Gao, Huang,
  Lv et~al.}]{yang2025qwen3}
An~Yang, Anfeng Li, Baosong Yang, Beichen Zhang, Binyuan Hui, Bo~Zheng, Bowen
  Yu, Chang Gao, Chengen Huang, Chenxu Lv, and 1 others. 2025.
\newblock Qwen3 technical report.
\newblock \emph{arXiv preprint arXiv:2505.09388}.

\bibitem[{Zhang et~al.(2025)Zhang, Khater, Rahli, Rebollo, and
  Crego}]{zhang2025systran}
Dakun Zhang, Yara Khater, Ramzi Rahli, Anna Rebollo, and Josep Crego. 2025.
\newblock \href {https://aclanthology.org/2025.wmt-1.35/} {{SYSTRAN} @ {WMT}
  2025 general translation task}.
\newblock In \emph{Proceedings of the Tenth Conference on Machine Translation},
  pages 599--606, Suzhou, China. Association for Computational Linguistics.

\bibitem[{Zheng et~al.(2023)Zheng, Chiang, Sheng, Zhuang, Wu, Zhuang, Lin, Li,
  Li, Xing, Zhang, Gonzalez, and Stoica}]{zheng2023judging}
Lianmin Zheng, Wei-Lin Chiang, Ying Sheng, Siyuan Zhuang, Zhanghao Wu, Yonghao
  Zhuang, Zi~Lin, Zhuohan Li, Dacheng Li, Eric~P. Xing, Hao Zhang, Joseph~E.
  Gonzalez, and Ion Stoica. 2023.
\newblock Judging {LLM}-as-a-judge with {MT-Bench} and chatbot arena.
\newblock In \emph{Advances in Neural Information Processing Systems
  (NeurIPS)}.

\bibitem[{Zheng et~al.(2025)Zheng, Li, Du, Qu, and Song}]{zheng2025hunyuan}
Mao Zheng, Zheng Li, Yang Du, Bingxin Qu, and Mingyang Song. 2025.
\newblock \href {https://aclanthology.org/2025.wmt-1.36/} {Shy-hunyuan-{MT} at
  {WMT}25 general machine translation shared task}.
\newblock In \emph{Proceedings of the Tenth Conference on Machine Translation},
  pages 607--613, Suzhou, China. Association for Computational Linguistics.

\end{thebibliography}

\end{document}